\documentclass[10pt,twocolumn,letterpaper]{article}

\usepackage{iccv}
\usepackage{xcolor}
\usepackage{times}
\usepackage{epsfig}
\usepackage{graphicx}
\usepackage{amsmath}
\usepackage{amssymb}
\usepackage{adjustbox}
\usepackage{tabularx}
\usepackage{algorithm}
\usepackage{algpseudocode}
\usepackage{multirow}
\usepackage{subfig}
\usepackage{comment}

\graphicspath{{}}
\newcolumntype{s}{D{.}{.}{1.2}}
\newcolumntype{d}{D{.}{.}{2.1}}

\newcolumntype{A}{>{\centering\arraybackslash}X}

\graphicspath{{}}
\def\MYFLAG{submit}
\ifdefined\MYFLAG

	\def\me{black}
\else

	\def\me{red}
\fi

\newcolumntype{b}{>{\hsize=2.3\hsize}X}
\newcolumntype{s}{>{\hsize=.45\hsize}X}
\newcolumntype{m}{>{\hsize=.9\hsize}X}

\usepackage[pagebackref=true,breaklinks=true,letterpaper=true,colorlinks,bookmarks=false]{hyperref}

\iccvfinalcopy 

\def\iccvPaperID{330} 

\DeclareMathOperator*{\argmax}{arg\,max}

\ificcvfinal\fi
\begin{document}

\title{Unpaired Image Captioning via Scene Graph Alignments}
\author{Jiuxiang Gu$^{1}$, Shafiq Joty$^{1,4}$, Jianfei Cai$^{1,2}$, Handong Zhao$^{3}$, Xu Yang$^1$, Gang Wang$^{5}$\\
	$^1$Nanyang Technological University, Singapore $^2$Monash University, Australia\\
	$^3$Adobe Research, USA $^4$Salesforce Research Asia, Singapore $^{5}$Alibaba Group, China \\
{\tt\small \{jgu004, srjoty, asjfcai, s170018\}@ntu.edu.sg, hazhao@adobe.com, gangwang6@gmail.com}}

\maketitle
\thispagestyle{empty}
\begin{abstract}
Most of current image captioning models heavily rely on paired image-caption datasets. However, getting large scale image-caption paired data is labor-intensive and time-consuming. In this paper, we present a scene graph-based approach for unpaired image captioning. Our framework comprises an image scene graph generator,  a sentence scene graph generator,  a scene graph encoder, and a sentence decoder. Specifically, we first train the scene graph encoder and the sentence decoder on the text modality. To align the scene graphs between images and sentences, we propose an unsupervised feature alignment method that maps the scene graph features from the image to the sentence modality. Experimental results show that our proposed model can generate quite promising results without using any image-caption training pairs, outperforming existing methods by a wide margin.
\end{abstract}

\section{Introduction}
Today's image captioning models heavily depend on paired image-caption datasets. Most of them employ an encoder-decoder framework~\cite{vinyals2017show,fang2015captions,gu2017stack,gu2017look,yang2019learning}, which uses a convolutional neural network (CNN)~\cite{gu2015recent} to encode an image into a feature vector and then a recurrent neural network (RNN) to decode it into a text sequence. However, it is worthwhile noticing that the overwhelming majority of image captioning studies are conducted in English~\cite{wiki:English_speaker}. The bottleneck is the lack of large scale image-caption paired datasets in other languages, and getting such paired data for each target language requires human expertise in a time-consuming and labor-intensive process.

\begin{figure}[t!]
	\centering
	\includegraphics[width=\linewidth]{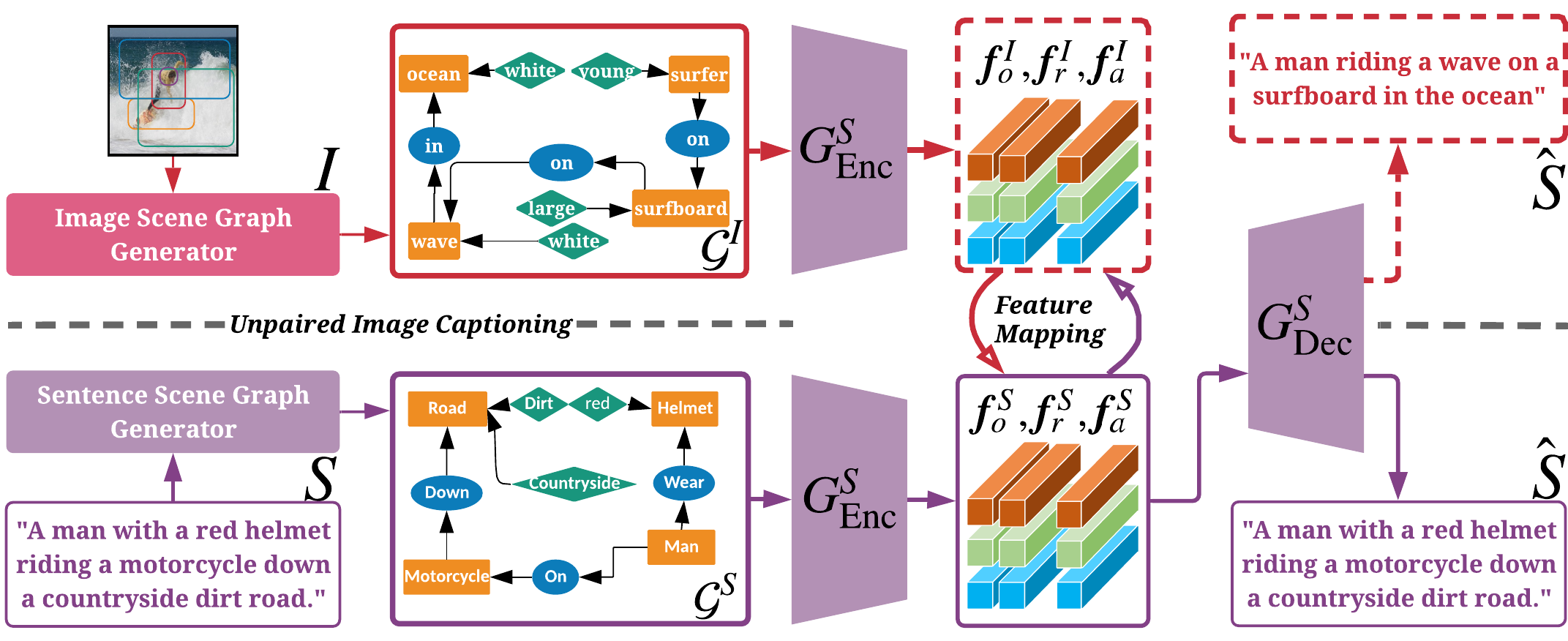}
    \vspace{-6mm}
	\caption{Illustration of our graph-based learning method. Our model consists of one visual scene graph detector (Top-Left), one fixed off-the-shelf scene graph language parser (Bottom-Left), a scene graph encoder $G_{\text{Enc}}^S$, a sentence decoder $G_{\text{Dec}}^S$, and a feature mapping module.}
	\label{fig:principle}
    \vspace{-3mm}
\end{figure}

Several encoder-decoder models have been proposed in recent years for unsupervised neural machine translation ~\cite{lample2017unsupervised,artetxe2017unsupervised}.
The key idea of these methods mainly relies on training denoising auto-encoders for language modeling and on sharing latent representations across the source and target languages for the encoder and the decoder. Despite the promising results achieved by the unsupervised neural machine translation, unpaired image-to-sentence translation is far from mature.

Recently, there have been few attempts at relaxing the requirement of paired image-caption data for this task. The first work in this direction is the pivot-based semi-supervised solution proposed by Gu~\etal~\cite{gu2018unpaired}, where they take a pivot language as a bridge to connect the source image and the target language caption. Their method requires an image-text paired data for the pivot language (Chinese), and a parallel corpus for the pivot to target translation. Feng~\etal~\cite{feng2018unsupervised} \textcolor{\me}{move} a step further, where they conduct purely unsupervised image captioning without relying on any labeled image-caption pairs. Their method uses a sentence discriminator along with a visual concept detector to connect the image and the text modalities through adversarial training. Although promising, the results of the existing methods are still far below compared to their paired counterparts.  

Unlike unsupervised neural machine translation where the encoders can be shared across the source and target languages, due to the different structures and characteristics of image and text modalities, the encoders of image and sentence cannot be shared to connect the two modalities. The critical challenge in unpaired image captioning is, therefore, the gap of information misalignment in images and sentences, so as to fit the encoder-decoder framework.

Fortunately, with recent breakthroughs in deep learning and image recognition, higher-level visual understanding tasks such as scene graph construction have become popular research topics with significant advancements~\cite{wacv18_zhao,ding2019boundary,ding2019semantic,ding2018context,aaai17_zhao,yang2018graph,gu2019graph}. \textcolor{\me}{Scene graph}, as an abstraction of objects and their complex relationships, provide rich semantic information of an image. The value of scene graph representation has been proven in a wide range of vision-language tasks, such as visual question answering~\cite{teney2017graph} and paired image captioning~\cite{yang2018graph}.

Considering the significant challenges that unpaired image captioning problem poses in terms of different characteristics between visual and textual modalities, in this paper, we propose a scene graph-based method that exploits the rich semantic information captured by scene graphs. Our framework comprises an image scene graph generator, a sentence scene graph generator, a scene graph encoder, a sentence decoder, and a feature alignment module that maps the features from image to sentence modality. Figure~\ref{fig:principle} sketches our solution. We first extract the sentence scene graphs from the sentence corpus and train the scene graph encoder and the sentence decoder on the text modality. To align the scene graphs between images and sentences, we use CycleGAN~\cite{zhu2017unpaired} to build the data correspondence between the two modalities. Specifically, given the unrelated image and sentence scene graphs, we first encode them with the scene graph encoder trained on the sentence corpus. Then, we perform unsupervised cross-modal mapping for feature level alignments with CycleGAN. By mapping the features, the encoded image scene graph is pushed close to the sentence modality, which is then used effectively as input to the sentence decoder to generate meaningful sentences.

The main contributions of this work include: (1) a novel scene graph-based framework for unpaired image captioning; (2) an unsupervised feature alignment method that learns the cross-modal mapping without any paired data. Our experimental results demonstrate the effectiveness of our proposed model in producing quite promising image captions. The comparison with recent unpaired image captioning methods validates the superiority of our method.

\section{Background}
\noindent\textbf{Paired Image Captioning.}
Image captioning has been extensively studied in the past few years. Most of the existing approaches are under the paired setting, that is, input images come with their corresponding ground-truth captions~\cite{vinyals2017show,hitschler2016multimodal,you2016image,gu2017stack,gu2017look,yang2018shuffle}.
One classic work in this setting is~\cite{vinyals2017show}, in which an image is encoded with a CNN, and the sentence is decoded with a Long Short-Term Memory (LSTM) network. Following this, many methods have been proposed to improve this encoder-decoder method. One of the notable improvements is the attention mechanism~\cite{xu2015show,gu2017stack,anderson2018bottom}, which allows the sentence decoder to dynamically focus on some related image regions during the caption generation process. Some other works explore other architectures for language modeling \cite{gu2017empirical,shuster2018engaging}. For example, Gu~\etal~\cite{gu2017empirical} \textcolor{\me}{introduce} a CNN-based language model for image captioning.
Another theme of improvements is to use reinforcement learning (RL) to address the exposure bias and loss-evaluation mismatch problems for sequence prediction~\cite{rennie2016self,gu2017stack}. The self-critical learning approach proposed in~\cite{rennie2016self} is a pioneering work, which well addresses the above two problems.

Our work in this paper is closely related to~\cite{yang2018graph}, which uses scene graph to connect images and sentences to incorporate inductive language bias. The key difference is that the framework in~\cite{yang2018graph} is based on the paired setting, while in this work, we learn the scene graph-based network under \textcolor{\me}{the} unsupervised training setting.

\noindent\textbf{Unpaired Image Captioning.}
More recently, some researchers started looking into the problem of image captioning in the unpaired setting~\cite{gu2018unpaired,feng2018unsupervised}, where there is no correspondence between images and sentences during training.
The first work on this task is the pivot-based solution proposed by Gu~\etal~\cite{gu2018unpaired}. In their setting, although they do not have any correspondence between images and sentences in the target language, they do require a paired image-caption dataset in the pivot language and another machine translation dataset which consists of sentences in the pivot language and the paired sentences in the target language. They connect \textcolor{\me}{the} pivot language sentences in different domains by shared word embeddings. The most recent work on unpaired image captioning is proposed by Fang~\etal~\cite{feng2018unsupervised}. They generate pseudo image-sentence pairs by feeding the visual concepts of images to a concept-to-sentence model and performing the alignment between image features and sentence features in an adversarial manner.

While several attempts have been made for the unpaired image captioning problem, this challenging task is far from mature. Arguably, compared to unpaired sentence-to-sentence ~\cite{lample2017unsupervised} and image-to-image~\cite{zhu2017unpaired} translations, unpaired image-to-sentence translation is more challenging because of the significantly different characteristics of the two modalities. In contrast to the existing unpaired image captioning methods~\cite{gu2018unpaired,feng2018unsupervised}, our proposed method adopts scene graph as an explicit representation to bridge the gap between image and sentence domains.

\section{Method}
In this section, we describe our unpaired image captioning framework. We first revisit the paired \textcolor{\me}{captioning} setting. 

\subsection{Paired Image Captioning Revisited}
In the paired captioning setting, our training goal is to generate a caption $S$ from an image $I$ such that $S$ is similar to its ground-truth caption. The popular encoder-decoder framework for image captioning can be formulated as:
\begin{align}
P(S|I)=\underbrace{P(V|I)}_{\text{Encoder}} \underbrace{P(S|V)}_{\text{Decoder}}\label{eq:basic_enc_dec}
\end{align}
where the encoder $P(V|I)$ encodes the image $I$ into the image features $V$ with a CNN model~\cite{gu2015recent}, and the decoder $P(S|V)$ predicts the image description $S$ from the image features $V$. The most common training objective is to maximize the probability of the ground-truth caption words given the image: $\sum_t \log p_{\theta_{I \rightarrow S}}(S_t|S_{0:t-1},I) $, where $p_{\theta_{I \rightarrow S}}(S_t|S_{0:t-1},I)$ corresponds to the Softmax output at time step $t$. During inference, the word $S_t$ is drawn from the dictionary $\mathcal{D}_S$ according to the Softmax distribution.

\subsection{Unpaired Image Captioning}
In the unpaired image captioning setting, we have a dataset of images {$\mathcal{I}=\{I_1, \ldots, I_{N_I}\}$}, and a dataset of sentences {$\mathcal{S}= \{S_1, \ldots, S_{N_S}\}$}, where $N_I$ and $N_S$ are the total numbers of images and sentences, respectively. In this setting, there is no alignment between $\mathcal{I}$ and $\mathcal{S}$. {In fact, $\mathcal{I}$ and $\mathcal{S}$ can be completely unrelated coming from two different domains.} Our goal is to train an image captioning model in a completely unsupervised way. In our setup, we assume that we have access to an off-the-shelf image scene graph detector and a sentence (or text) scene graph parser.

As shown in Figure~\ref{fig:principle}, {our proposed image captioning model consists of one image scene graph generator, one sentence scene graph generator, one scene graph encoder $G_{\text{Enc}}^S$, one attention-based decoder for sentence generation $G^S_{\text{Dec}}$, and a cycle-consistent feature alignment module}. {Given an image $I$ as input, our method first extracts an image scene graph $\mathcal{G}^I$ using the scene graph generator. It then maps $\mathcal{G}^I$ to the sentence scene graph $\mathcal{G}^S$ from which the \textcolor{\me}{RNN-based} decoder generates a sentence $S$.} More formally, the image captioner $P(S|I)$ in the unpaired setting can be decomposed into the following submodels,
\begin{align}
 I &\rightarrow~~\mathcal{G}^I \\
 P(S|I) &= \underbrace{P(\mathcal{G}^S|\mathcal{G}^I)}_{\text{Unpaired Mapping}} \underbrace{P(S|\mathcal{G}^S)}_{\text{Decoder}}\label{eq:graph_enc_dec}
\end{align}
where $\mathcal{G}^I$ and $\mathcal{G}^S$ are the image scene graph and the sentence scene graph, respectively. The most crucial component in Eq.~\eqref{eq:graph_enc_dec} is the unpaired mapping of image and text scene graphs. In our approach, this mapping is done in the feature space. In particular, we encode the image and sentence scene graphs into feature vectors and learn to map the feature vectors across the two modalities. We reformulate Eq.~\eqref{eq:graph_enc_dec} as follows:  
\begin{eqnarray}\label{eq:graph_enc_dec_re}
P(S|I) &=& P(\mathcal{G}^S|\mathcal{G}^I) P(S|\mathcal{G}^S) \nonumber \\ 
&\approx& P(\boldsymbol{f}^I|\mathcal{G}^I)P(\boldsymbol{f}^S|\boldsymbol{f}^I) P(S|\boldsymbol{f}^S)
\end{eqnarray}
where $P(\boldsymbol{f}^I|\mathcal{G}^I)$ is a graph encoder, $P(S|\boldsymbol{f}^S)$ is an RNN-based sentence decoder, and $P(\boldsymbol{f}^S|\boldsymbol{f}^I)$ is a cross-modal feature mapper in the unpaired setting.
In our implementation, we learn the scene graph encoder and the RNN-based decoder on the text modality first, and then we try to map the image scene graph into a common feature space (\ie, the text space) so that the same sentence decoder can be used to decode the sentence from the mapped image features.  

The sentence encoding and decoding {processes} can be formulated as the following two steps: 
\begin{align}
 &S \rightarrow ~~\mathcal{G}^S \\
 &\hat{S} = \argmax_S P(S|\boldsymbol{f}^S)P(\boldsymbol{f}^S|\mathcal{G}^S) \label{eq:sentgraph}
\end{align}
where $\hat{S}$ is the reconstructed sentence. We train the model to {enforce} $\hat{S}$ to be close to the original sentence $S$.

In the following, we describe the scene graph generator in Sec.~\ref{sec:sg_gen}, the scene graph encoder in Sec.~\ref{sec:sg_enc}, the sentence decoder in Sec.~\ref{sec:sent_dec}, and our unpaired feature mapping process in Sec.~\ref{sec:gan_learn}.

\subsubsection{Scene Graph Generator}\label{sec:sg_gen}
Formally, a scene graph is a graph $\mathcal{G}=(\mathcal{V},\mathcal{E})$ containing a set of nodes $\mathcal{V}$ and a set of edges $\mathcal{E}$. As exemplified in Figure \ref{fig:principle}, the nodes can be of three types: object node, attribute node, and relationship node. We denote $o_i$ as the $i$-th object, $r_{i,j}$ as the relation between object $o_i$ and $o_j$, and $a_{i}^{l}$ as the $l$-th attribute of object $o_i$.

An image scene graph generator contains an object detector, an attribute classifier, and a relationship classifier. We use Faster-RCNN~\cite{ren2015faster} as the object detector, MOTIFS ~\cite{zellers2018neural} as the relationship detector, and {an additional classifier for attribute identification~\cite{yang2018graph}.}

To generate the sentence scene graph $\mathcal{G}^S$ for a sentence, we first parse the sentence into a syntactic tree using the parser provided by~\cite{anderson2016spice}, which uses a syntactic dependency tree  built by~\cite{klein2003accurate}. Then, we transform the tree into a scene graph with a rule-based method~\cite{schuster2015generating}.

\subsubsection{Scene Graph Encoder}\label{sec:sg_enc}
We follow \cite{yang2018graph} to encode a scene graph. Specifically, {we represent each node as a $d_e$-dimensional feature vector, and use three different spatial graph convolutional encoders to encode the three kinds of nodes by considering their neighborhood information in the scene graph.}

\vspace{-1em}
\paragraph{Encoding objects.}

In a scene graph (image or sentence), an object $o_i$ can play {either} a \emph{subject} or an \emph{object} role in a relation triplet depending on the direction of the edge. Therefore, for encoding objects, we consider what relations they are associated with and what roles they play in that relation. Let $\langle o_i,o_j,r_{i,j} \rangle$ denote the triplet for relation $r_{i,j}$, where $o_i$ plays a \emph{subject} role and $o_j$ plays an \emph{object} role. The encoding for object $o_i$, {that is {$\boldsymbol{x}_{o_i}\in \mathbb{R}^{d_x}$} is computed by}
\begin{align}
    \boldsymbol{x}_{o_i}=&\frac{1}{N_{r_i}}\sum_{o_j}g_s(\boldsymbol{e}_{o_i},\boldsymbol{e}_{o_j},\boldsymbol{e}_{r_{i,j}})\nonumber\\ +&\frac{1}{N_{r_i}}\sum_{o_k}g_o(\boldsymbol{e}_{o_k},\boldsymbol{e}_{o_i},\boldsymbol{e}_{r_{k,i}}) \label{eq:obj}
\end{align}
where $\boldsymbol{e}_{o_i}\in \mathbb{R}^{d_e}$ and $\boldsymbol{e}_{r_{i,j}}\in \mathbb{R}^{d_e}$ are the embeddings (randomly initialized) representing the object ${o_i}$ and the relation $r_{i,j}$, respectively; $g_s(\cdot)$ and $g_o(\cdot)$ are the spatial graph convolution operations for objects as a  \emph{subject} and as an \emph{object}, respectively; and $N_{r_i}$ is the total number of relation triplets that $o_i$ is associated with in the scene graph. 

\vspace{-1em}
\paragraph{{Encoding attributes.}}
An object $o_i$ may have multiple attributes in the scene graph. The encoding of an object based on its attributes, \ie, {$\boldsymbol{x}_{a_i}\in \mathbb{R}^{d_x}$} is computed by:
\begin{align}
    \boldsymbol{x}_{a_i} =\frac{1}{N_{a_i}}\sum_{l}g_a(\boldsymbol{e}_{o_i},\boldsymbol{e}_{a_i^l})
\end{align}
where $N_{a_i}$ is the total number of attributes that object $o_i$ has, and $g_a(\cdot)$ is the spatial convolutional operation for attribute based encoding.

\vspace{-1em}
\paragraph{Encoding relations.}
Each relation $r_{i,j}$ is encoded into {$\boldsymbol{x}_{r_{i,j}} \in \mathbb{R}^{d_x}$} by considering the objects that the relation connects in the relation triplet,
\begin{align}
    \boldsymbol{x}_{r_{i,j}} = g_r(\boldsymbol{e}_{o_i}, \boldsymbol{e}_{o_j}, \boldsymbol{e}_{r_{i,j}})
\end{align}
where $g_r(\cdot)$ is the associated convolutional operation. 

\begin{figure}[t!]
	\centering
	\includegraphics[width=\linewidth]{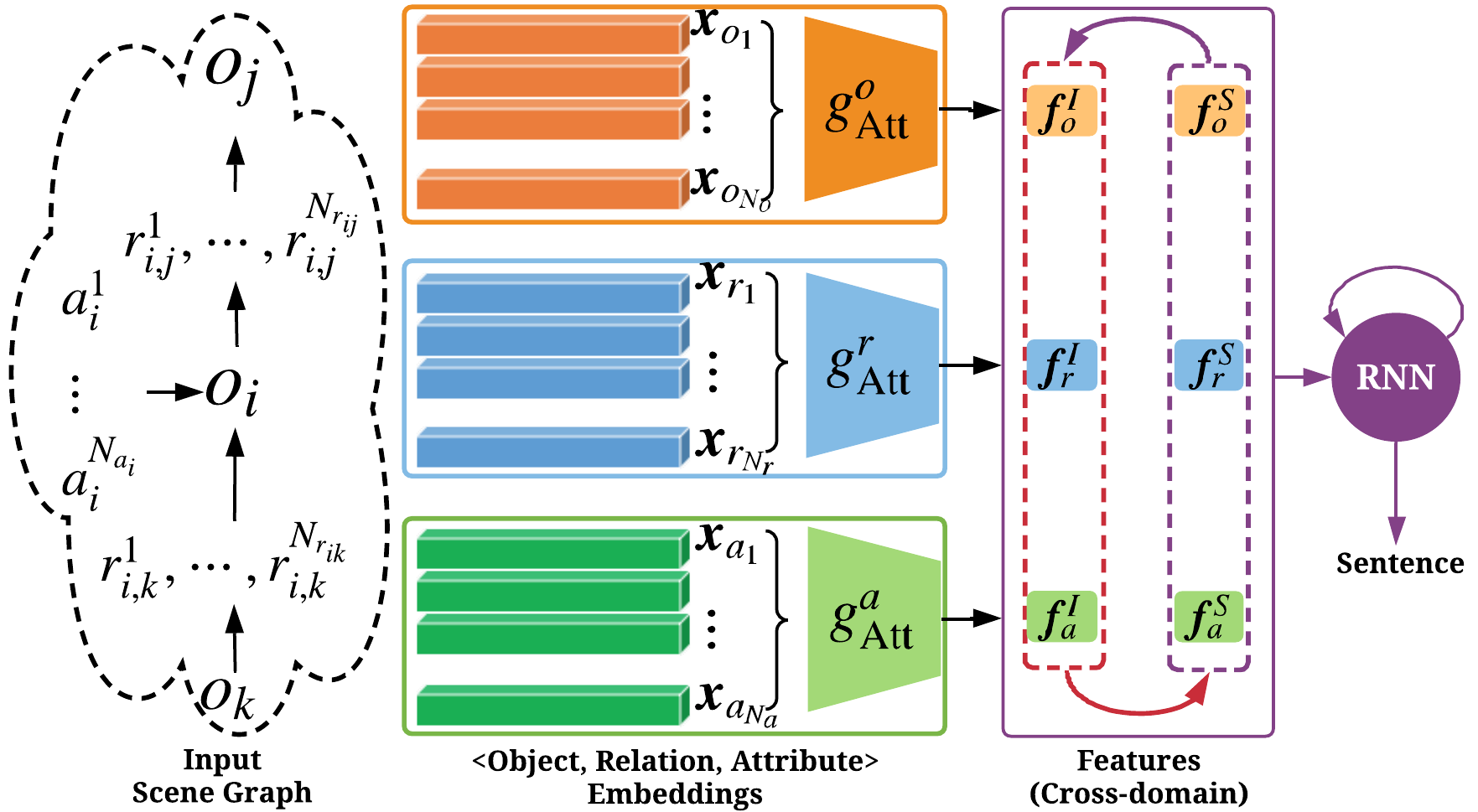}
    \vspace{-6mm}
	\caption{The architectures for scene graph encoding, attention, and sentence decoding; { $g_{\text{Att}}^o$, $g_{\text{Att}}^r$, and $g_{\text{Att}}^a$ are attention modules for each kind of features, respectively}.}
	\label{fig:sg_enc_dec}
    \vspace{-3mm}
\end{figure}

After graph encoding, for each image scene graph {or} sentence scene graph, we have three sets of embeddings:
{
\begin{minipage}[t]{\linewidth}
\begin{equation}
\begin{aligned}
    \mathcal{X}^k_o=&[\boldsymbol{x}^k_{o_1},\cdots, \boldsymbol{x}^k_{o_{N_o^k}}]\\ \mathcal{X}^k_r=&[\boldsymbol{x}^k_{r_1},\cdots, \boldsymbol{x}^k_{r_{N_r^k}}]\\ \mathcal{X}^k_a=&[\boldsymbol{x}^k_{a_1},\cdots, \boldsymbol{x}^k_{a_{N_a^k}}]\end{aligned},\quad k\in\{I,S\}
\end{equation}
\end{minipage}
}\\
where $N_o^k$, $N_r^k$, and $N_a^k$ can be  different from each other. Figure~\ref{fig:sg_enc_dec} illustrates the encoding process.

\subsubsection{Sentence Decoder}\label{sec:sent_dec}
The goal of the sentence decoder is to generate a sentence {$\hat{S}$} from the encoded embeddings, $\mathcal{X}^k_o$, $\mathcal{X}^k_r$, and $\mathcal{X}^k_a$. However, these three sets of embeddings are of different lengths and contain different information. Therefore, their importance for the sentence decoding task also vary. In order to compute a relevant context for the decoding task effectively, we use three attention modules, one for each type of embeddings. {The attention module $g_{\text{Att}}^o$ over $\mathcal{X}^k_o$ is defined as}:
\begin{align}
    \boldsymbol{f}_o^k=\sum_{i=1}^{N_o^k} \alpha_i \boldsymbol{x}_{o_i}^k; \hspace{1em} 
    \alpha_i =\frac{\exp(\boldsymbol{w}_o^T \boldsymbol{x}_{o_i}^k)}{\sum_m \exp(\boldsymbol{w}_o^T \boldsymbol{x}_{o_m}^k)}
\end{align}
where $\boldsymbol{w}_o$ is the associated (learnable) weight vector. The attentions over $\mathcal{X}^k_r$, and $\mathcal{X}^k_a$ are similarly defined to get the respective attention vectors, {$\boldsymbol{f}_r^k\in \mathbb{R}^{d_f}$ and $\boldsymbol{f}_a^k\in \mathbb{R}^{d_f}$}. 

The attention vectors are then combined to get a triplet level embedding, which is then fed into an {RNN-based decoder} to generate the sentence {$\hat{S}$}. The following sequence of operations formally describes the process. 
\begin{align}
\boldsymbol{f}^k_{\text{ora}} &=  g_{\text{ora}}([\boldsymbol{f}_o^k, \boldsymbol{f}_r^k, \boldsymbol{f}_a^k])\label{eq:triplet_emb}\\
\mathbf{o}_t, \mathbf{h}_t &=   \text{RNN}_{\text{Dec}}(\boldsymbol{f}^k_{\text{ora}}, \mathbf{h}_{t-1}, \hat{S}_{t-1})\\
\hat{S}_t &\approx \text{softmax}(\mathbf{W}_o \mathbf{o}_t)
\end{align}
where $g_{\text{ora}}(\cdot)$ is a neural network that generates a triplet level embedding, and $\mathbf{o}_t$ is the cell output of the decoder at time step $t$.

\subsubsection{Training and Inference}\label{sec:gan_learn}
We first train the graph encoder and the sentence decoder in the text modality (Eq. \eqref{eq:sentgraph}), and then perform a feature level alignments for cross-modal unsupervised mapping.

\noindent\textbf{Training in Text Modality.} The graph convolutional encoding of a sentence scene graph $\mathcal{G}^S$ into a feature representation $\boldsymbol{f}^S_{\text{ora}}$, and reconstructing the original sentence $S$ from it {are shown at} the bottom part of Figure~\ref{fig:principle}, where the encoder and the decoder are denoted as $G_{\text{Enc}}^S$ and $G_{\text{Dec}}^S$, respectively. We first train  $G_{\text{Enc}}^S$ and $G_{\text{Dec}}^S$ models by minimizing the cross-entropy (XE) loss:
\begin{equation}
    \mathcal{L}_{\text{XE}} (\theta_{\mathcal{G} \rightarrow S}) = - \sum_t \log p_{\theta_{\mathcal{G} \rightarrow S}}(S_t|S_{0:t-1}) \label{equ:equ_celoss}
\end{equation}
where $\theta_{\mathcal{G} \rightarrow S}$ are the parameters of $G_{\text{Enc}}^S$ and $G_{\text{Dec}}^S$, $ p_{\theta_{\mathcal{G} \rightarrow S}}(S_t|S_{0:t-1})$ is the output probability of $t$-th word in the sentence given by the sentence decoder.

We further employ a reinforcement learning (RL) loss that takes the entire sequence into account. Specifically, we take the CIDEr~\cite{vedantam2015cider} score as the reward and optimize $\theta_{\mathcal{G} \rightarrow S}$ by minimizing the negative expected rewards as follows:
\begin{equation}
    \mathcal{L}_{\text{RL}} (\theta_{\mathcal{G} \rightarrow S}) = -\mathbb{E}_{\tilde{S} \sim P_{\theta_{\mathcal{G} \rightarrow S}}}[r(\tilde{S})]
    \label{equ:equ_rlloss}
\end{equation}
where $r(\tilde{S})$ is the reward calculated by comparing the sampled sentence $\tilde{S}$ with the ground-truth sentence ${S}$ using the CIDEr metric. In our model, we follow the RL approach proposed in~\cite{rennie2016self,gu2017stack}.

\begin{figure}[t!]
	\centering
	\includegraphics[width=\linewidth]{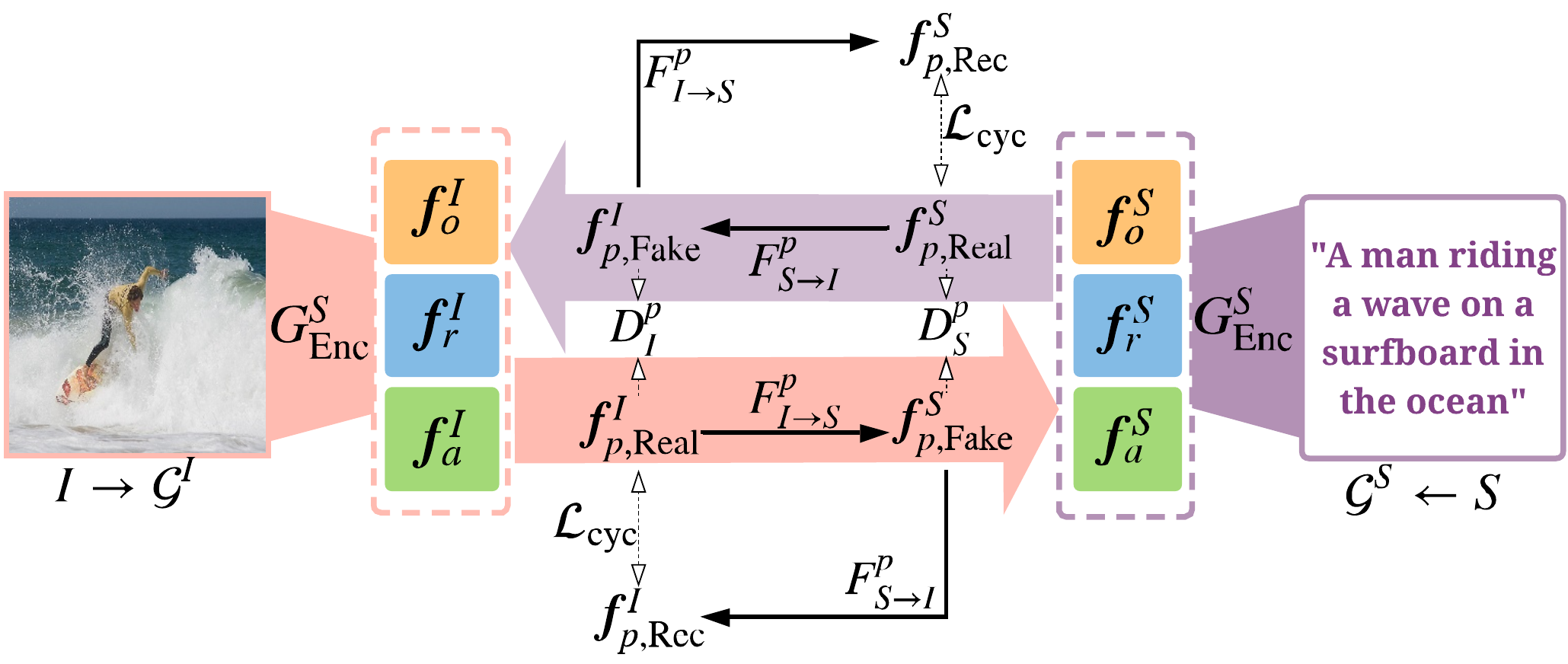}
    \vspace{-6mm}
	\caption{Conceptual illustration of our unpaired feature mapping. {For each kind of embedding $p$, there are two mapping functions $F^p_{I\rightarrow S}$ and $F^p_{S\rightarrow I}$, and two associated adversarial discriminators $D^p_I$ and $D^p_S$}.}
	\label{fig:cyclegan}
    \vspace{-3mm}
\end{figure}

\noindent\textbf{Unsupervised Mapping of Scene Graph Features.} To adapt the learned model from sentence modality to the image modality, we need to translate the scene graph from the image to the sentence modality.  We take the discrepancy in the modality of scene graphs directly into account by aligning the representation of the image scene graph with the sentence scene graph.
We propose to use CycleGAN~\cite{zhu2017unpaired} to learn the feature alignment across domains.

Figure~\ref{fig:cyclegan} illustrates our idea. {Given two sets of unpaired features $\boldsymbol{f}_p^I$ and $\boldsymbol{f}_p^S$, where $p\in \{o, r, a\}$, we have two mapping functions $F_{I \rightarrow S}^p(\cdot)$ and $F_{S \rightarrow I}^p(\cdot)$, and two discriminators $D_{S}^p$ and $D_{I}^p$. {$F^p_{I \rightarrow S}(\cdot)$} maps the image features to the sentence features, and $F^p_{S \rightarrow I}(\cdot)$ maps the sentence features to the image features}. The discriminators are trained to distinguish the \emph{real} (original modality) features from the \emph{fake} (mapped) features. The mappers are trained to fool the respective discriminators through adversarial training. 

For image to text mapping, the adversarial loss is
\begin{align}
    \mathcal{L}_{\text{GAN}}(F^p_{I\rightarrow S},D^p_S)= \mathbb{E}_{S}[\log D^p_S(\boldsymbol{f}_{p,\text{Real}}^S)]&\nonumber \\
    + \mathbb{E}_{I}[\log (1-D^p_S(F^p_{I \rightarrow S}(\boldsymbol{f}_{p,\text{Real}}^I))]&\label{eq:gan_i_s}
\end{align}
Similarly, for sentence to image mapping, we have the similar adversarial loss, 
\begin{align}
    \mathcal{L}_{\text{GAN}}(F^p_{S\rightarrow I},D^p_I)=\mathbb{E}_{I}[\log D^p_I(\boldsymbol{f}_{p,\text{Real}}^I)]&\nonumber \\
    +\mathbb{E}_{S}[\log (1-D^p_I(F^p_{S\rightarrow I}(\boldsymbol{f}_{p,\text{Real}}^S))]&\label{eq:gan_s_i}
\end{align}

Due to the unpaired setting, the mapping from the source to the target modality is highly under-constrained. To make the mapping functions cycle-consistent, CycleGAN introduces a cycle consistency loss to regularize the training,
\begin{align}
    \mathcal{L}_{\text{cyc}}(F^p_{S\rightarrow I},F^p_{I \rightarrow S})=&\mathbb{E}_{I}[\| \boldsymbol{f}_{p,\text{Rec}}^I-\boldsymbol{f}_{p,\text{Real}}^I \|_1]\nonumber\\
    +&\mathbb{E}_{S}[\|\boldsymbol{f}_{p,\text{Rec}}^S-\boldsymbol{f}_{p,\text{Real}}^S \|_1]
\end{align}
where $\boldsymbol{f}_{p,\text{Rec}}^I$ and $\boldsymbol{f}_{p,\text{Rec}}^S$ are the reconstructed features in the image and text modalities, respectively.

Formally, our overall objective for unpaired feature mapping is to optimize the following loss:
\begin{align}
    \mathcal{L}_{\text{Adv}}(\theta_{I \leftrightarrow S})&=\sum_{p\in \{o, r, a\}} \mathcal{L}_{\text{Adv}}(\theta^p_{I \leftrightarrow S})\\
    \mathcal{L}_{\text{Adv}}(\theta^p_{I \leftrightarrow S})&=\mathcal{L}_{\text{GAN}}(F^p_{S\rightarrow I},D^p_I)+\mathcal{L}_{\text{GAN}}(F^p_{I\rightarrow S},D^p_S)\nonumber\\
    &+\lambda \mathcal{L}_{\text{cyc}}(F^p_{S\rightarrow I},F^p_{I \rightarrow S})
    \label{eq:cycle_gan_final}
\end{align}
where $\theta^p_{I \leftrightarrow S}$ are the parameters of the two mapping functions and the discriminators for {each kind of embedding $p$}, and $\lambda$ is a hyperparameter to control the regularization.

\noindent\textbf{Cross-modal Inference.} 
During inference, given an image $I$,  we generate its corresponding scene graph $\mathcal{G}^I$ using a pre-trained image scene graph generator, use the scene graph encoder to get the image features $\boldsymbol{f}^I_p$, which are then mapped through the image-to-text mapper $F^p_{I\rightarrow S}$. The mapped features are then used for sentence generation using the sentence decoder. The cross-modal inference process can be formally expressed as:  
\begin{align}
    \hat{S}=&\arg\max_S P(S|\boldsymbol{f}_{\text{ora}}^I)P(\boldsymbol{f}_r^I,\boldsymbol{f}_o^I,\boldsymbol{f}_a^I|\mathcal{G}^I)\\
    \boldsymbol{f}_{\text{ora}}^I=&g_{\text{ora}}([F^o_{I\rightarrow S}(\boldsymbol{f}_{\text{o}}^I),F^r_{I\rightarrow S}(\boldsymbol{f}_{\text{r}}^I),F^a_{I\rightarrow S}(\boldsymbol{f}_{\text{a}}^I)])
\end{align}
where $g_{\text{ora}}(\cdot)$ is the same module as Eq.~\eqref{eq:triplet_emb}.

\section{Experiments}
In this section, we evaluate the effectiveness of our proposed method.  We first introduce the  datasets and the experimental settings. Then, we present the performance comparisons as well as ablation studies to understand the impact of different components of our framework.

\subsection{Datasets and Setting}
Table~\ref{tab:dataset} shows the statistics of the training datasets used in our experiments. We use Visual Genome (VG) dataset \cite{krishna2017visual} to train our image scene graph generator. We filter the object, attribute, and relation annotations by keeping those that appear more than 2,000 times in the training set. The resulting dataset contains 305 objects, 103 attributes, and 64 relations (a total of 472 items).  

We collect the image descriptions from the training split of MSCOCO~\cite{lin2014microsoft} and use them as our sentence corpus to train the scene graph encoder and the sentence decoder. In pre-processing, we tokenize the sentences and {convert all the} tokens to lowercase. The tokens that appear less than five times are treated as $\langle$UNK$\rangle$ tokens. The maximum caption length is fixed to 16, and all the captions longer than 16 are truncated. This results in a base vocabulary of 9,487 words.
For sentence scene graph generation, we generate the scene graph using the language parser in \cite{anderson2016spice,yang2018graph}. We perform a filtering process by removing objects, relations, and attributes which appear less than 10 times in all the parsed scene graphs.
{After this filtering, we obtain 5,364 objects, 1,308 relations, and 3,430 attributes. This gives an extended vocabulary where the previous 9,487 words are consistent with the base vocabulary. The embeddings for the vocabulary items are randomly initialized.}

\begin{table}[ht]
\vspace{-3mm}
\caption{Statistics of the training datasets.}
\vspace{-6mm}
\begin{center}
\setlength{\tabcolsep}{4pt}
\begin{tabular}{l|r|r|r}
\hline
\multirow{2}*{{Scene Graph}} & \multicolumn{3}{c}{Vocabulary Size} \\
\cline{2-4}
& {\#Object} &{\#Attribute}& {\#Relation}\\
\hline
Image (VG) & 305 &103 & 64 \\
\hline
Sentence (MSCOCO)&5,364&3,430 &1,308\\
\hline
\end{tabular}
\end{center}
\label{tab:dataset}
\vspace{-6mm}
\end{table}

For learning the mapping between the modalities, the unpaired training data is intentionally collected by shuffling the images and the sentences from MSCOCO randomly. We validate the effectiveness our method on the same test splits as used in~\cite{gu2018unpaired,feng2018unsupervised} for a fair comparison. The widely used CIDEr-D~\cite{vedantam2015cider}, BLEU~\cite{papineni2002bleu}, METEOR~\cite{banerjee2005meteor}, and SPICE~\cite{anderson2016spice} are used to measure the quality of the generated captions.

\begin{figure*}[ht!]
\vspace{-3mm}
\subfloat[Object Features (Raw)]{\label{fig:pre_tsne_obj}\includegraphics[width=0.25\linewidth]{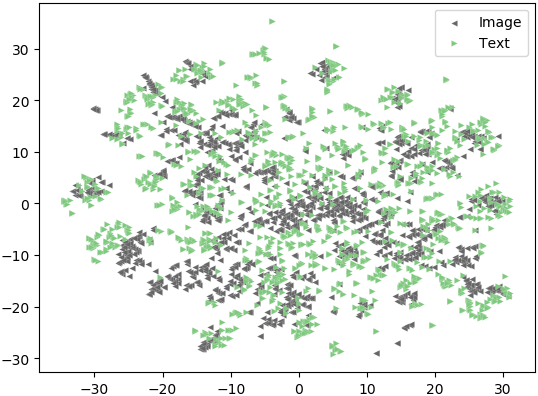}}
\subfloat[Relation Features (Raw)]{\label{fig:pre_tsne_rel}\includegraphics[width=0.25\linewidth]{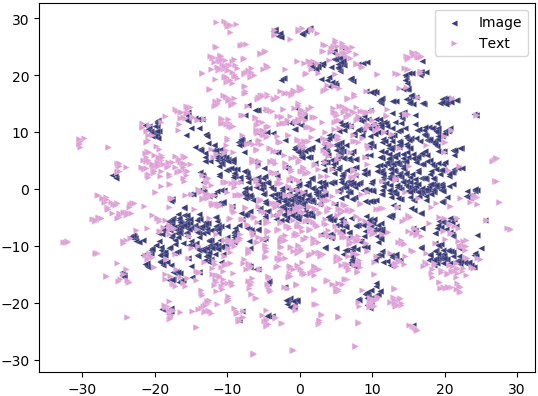}}%
\subfloat[Attribute Features (Raw)]{\label{fig:pre_tsne_atr}\includegraphics[width=0.25\linewidth]{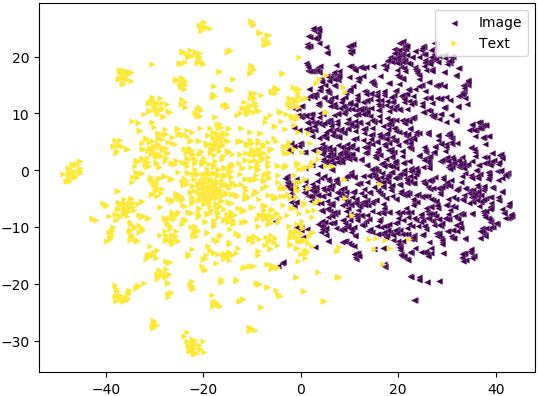}}
\subfloat[Triplet Features (Raw)]{\label{fig:pre_tsne_N_map}\includegraphics[width=0.25\linewidth]{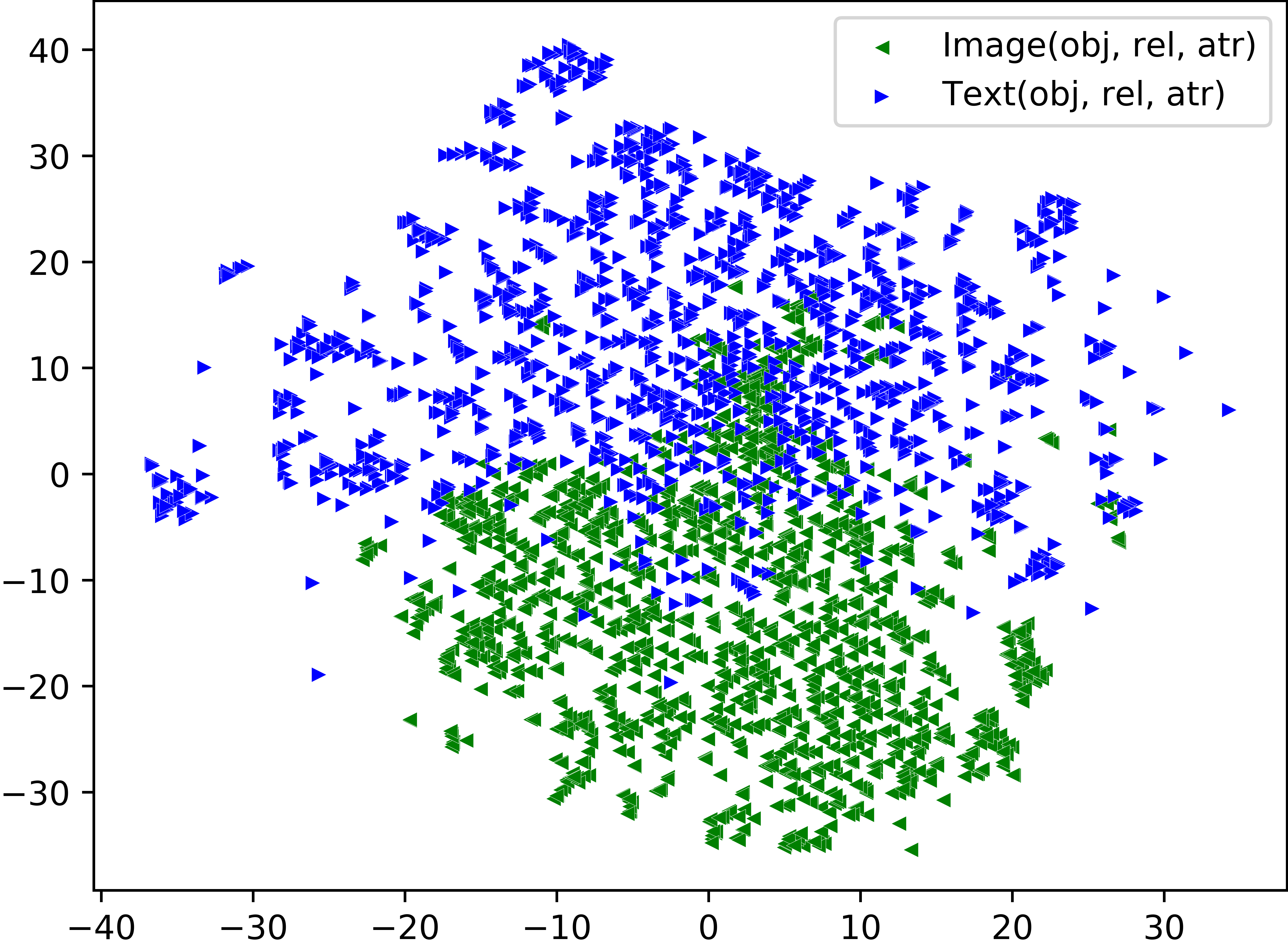}}\\[-2ex]
\subfloat[Object Features (Aligned)]{\label{fig:pre_tsne_obj_gan}\includegraphics[width=0.25\linewidth]{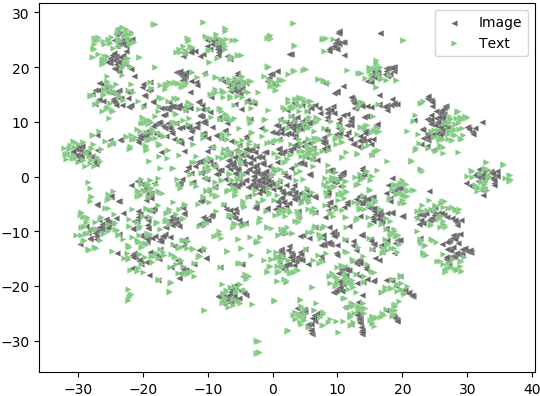}}
\subfloat[Relation Features (Aligned)]{\label{fig:pre_tsne_rel_gan}\includegraphics[width=0.25\linewidth]{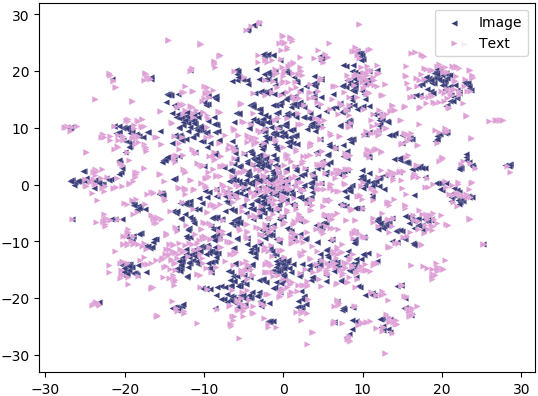}}%
\subfloat[Attribute Features (Aligned)]{\label{fig:pre_tsne_atr_gan}\includegraphics[width=0.25\linewidth]{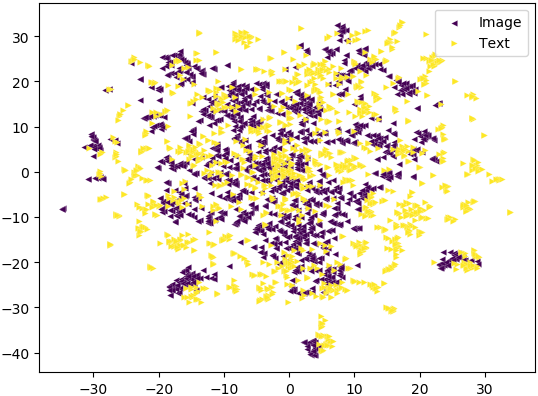}}
\subfloat[Triplet Features  (Aligned)]{\label{fig:pre_tsne_Y_map}\includegraphics[width=0.25\linewidth]{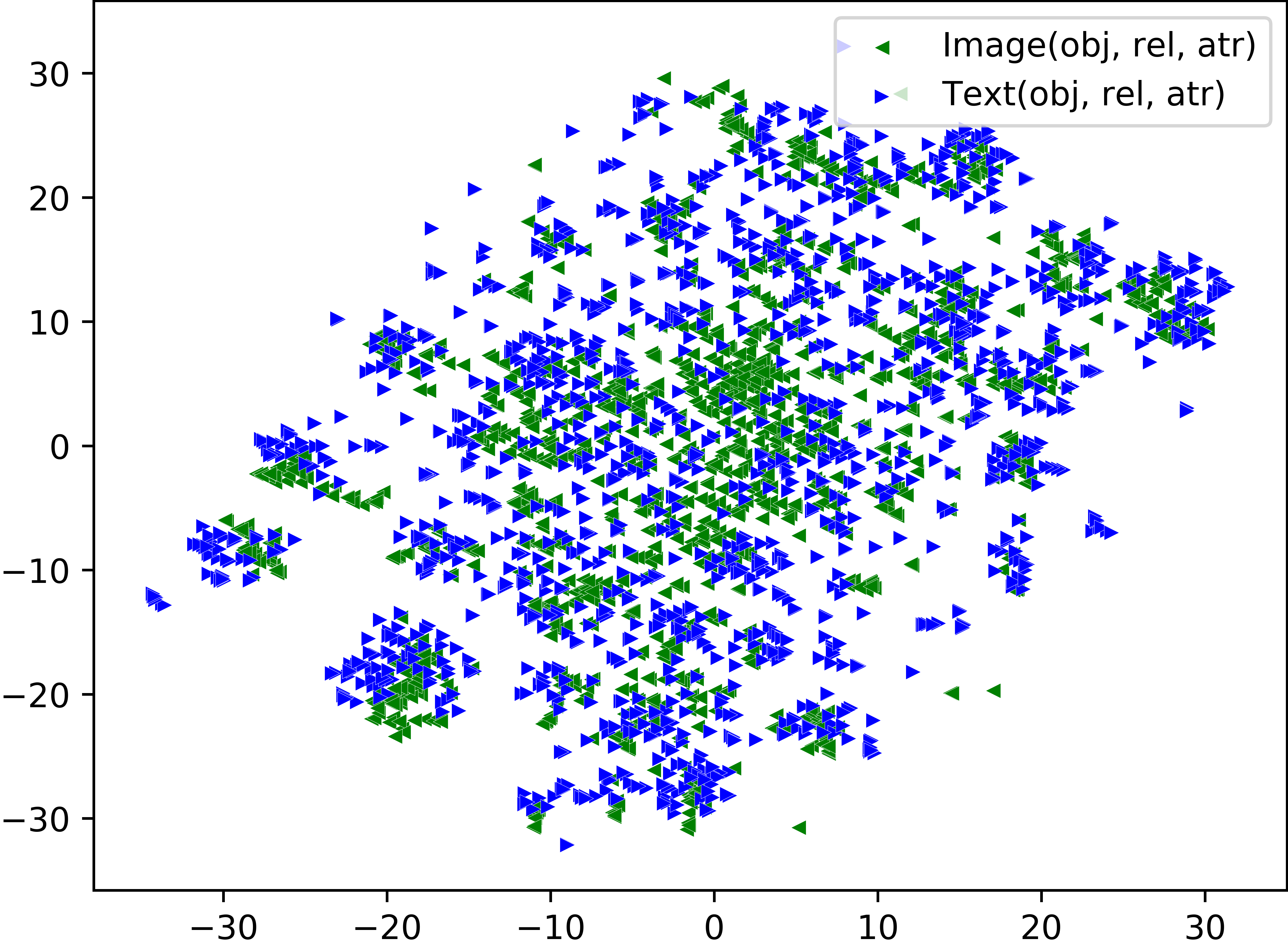}}
\vspace{-3mm}
\caption{Visualization of features in 2D space by t-SNE~\cite{maaten2008visualizing}. We plot the scatter diagrams for 1,500 samples.}
\label{fig:vis_feature}
\vspace{-3mm}
\end{figure*}

\subsection{Implementation Details}

We follow \cite{yang2018graph} to train our image scene graph generator on VG. We first train a Faster-RCNN and use it to \textcolor{\me}{identify} the objects in each image. We select at least 10 and at most 100 objects for an image.  The object features extracted by RoI pooling are used as input to the object detector, the relation classifier, and the attribute classifier. We adopt the LSTM-based relation classifier from~\cite{zellers2018neural}. Our attribute classifier is a single hidden layer network with ReLU activation (\ie,  fc-ReLU-fc-Softmax), and we keep only the three most probable attributes for each object.
{For scene graph encoding, we set $d_e=d_x=d_f=1000$.}
We implement $g_s$, $g_o$, $g_r$, $g_a$, and $g_{\text{ora}}$ (Eq. \eqref{eq:obj} - \eqref{eq:triplet_emb}) as fully-connected layers with ReLU activations. The two mapping functions in Eq.~\eqref{eq:gan_i_s} and Eq.~\eqref{eq:gan_s_i} are implemented as fully-connected layers with leaky ReLU activations. 

The sentence decoder has two LSTM layers.  The input to the first LSTM is the word embeddings and its previous hidden state. The input to the second LSTM is the concatenation of three terms: the triplet embedding  $\boldsymbol{f}_{\text{ora}}$, the output from the first LSTM, and its previous hidden state. We set the number of hidden units in each LSTM to 1,000.

During training, we first train the network with the cross-entropy loss (Eq.~\eqref{equ:equ_celoss}) for $20$ epochs and then fine-tune it with RL loss in Eq.~\eqref{equ:equ_rlloss}. The learning rate is initialized to $4\times 10^{-4}$ for all parameters and decayed by $0.8$ after every $5$ epoch. We use Adam ~\cite{kingma2014adam} for optimization with a batch size of $50$. During the (unpaired) alignment learning, we freeze the parameters of the scene graph encoder and the sentence decoder, and only learn the {mapping functions} and the discriminators. \textcolor{\me}{For all the experiments, we empirically set $\lambda$ to 10 in Eq.~\eqref{eq:cycle_gan_final}}. During inference, we use beam search with a beam size of 5.  

For quantifying the efficacy of the proposed framework, we use several baselines for performance comparison.

\noindent\textbf{Graph-Enc-Dec (Avg)}. This baseline learns the graph encoder $G_{\text{Enc}}^S$ and the sentence decoder $G_{\text{Dec}}^S$ only on sentence corpus. It takes the average operation (as opposed to attention) over the three sets of features: $\mathcal{X}_o^k$, $\mathcal{X}_r^k$, and $\mathcal{X}_a^k$. During testing, we directly feed the image scene graph $\mathcal{G}^I$ to this model and get the image description.

\noindent\textbf{Graph-Enc-Dec (Att$^{\ast}$)}. This model shares the same setting with Graph-Enc-Dec (Avg) but replaces the average operation with a shared attention mechanism for all three sets (i.e., same attention for object, attribution, and relation).

\noindent\textbf{Graph-Enc-Dec (Att)}. This model modifies the Graph-Enc-Dec (Att$^{\ast}$) with an independent attention mechanism for each set of features.

\noindent\textbf{Graph-Align}. This is our final model. It is initialized with the trained parameters from Graph-Enc-Dec (Att) that uses separate attentions, and then it also learns the feature mapping functions using adversarial training.

\subsection{Quantitative Results}
\noindent{\textbf{Investigation on Sentence Decoding.}}
In this experiment, we first train the network with Eq.~\eqref{equ:equ_celoss}, and then fine-tune it with Eq.~\eqref{equ:equ_rlloss} on the sentence corpus. Table~\ref{tab:txt_ssg_decoder} compares the results of three baseline models on the sentence corpus. It can be seen that the attention-based model performs better than the average-based model in all metrics, which demonstrates that weighting over features can better model the global dependency of features. Note that separate attention model for each set of features can significantly improve the performance. {The inconsistent alignment of three kinds of features in Figure~\ref{fig:vis_feature} also supports that we should treat these sets of features separately}.

\begin{table}[ht]
\vspace{-3mm}
\caption{Results for different sentence scene graph decoders on MSCOCO test split, where B@n refers to BLEU-n, M refers to METEOR, and C refers to CIDEr. All values are reported in percentage (bold numbers are the best results).}
\vspace{-6mm}
\begin{center}
\small
\setlength{\tabcolsep}{3pt}
\begin{tabular}{l| c |c |c |c |c |c }
\hline
{Methods}  & {{B@1}} & {{B@2}} & {{B@3}} & {{B@4}} & {{M}} & {\textbf{C}} \\
\hline
Graph-Enc-Dec(Avg) & 84.3 & 71.8 & 58.8 & 47.1 & 31.0 & 129.4\\
\hline
Graph-Enc-Dec(Att$^{\ast}$) & 91.8 & 80.3 & 67.5 & 55.5 & 34.3 & 151.4\\
\hline
Graph-Enc-Dec(Att) & 94.1 & 84.6 & 72.9 & 61.5 & 36.3 & 168.8\\
\hline
\end{tabular}
\end{center}\label{tab:txt_ssg_decoder}
\vspace{-6mm}
\end{table}

\begin{table}[ht]
\vspace{-3mm}
\caption{Results for different baselines without GAN training on the test split of the MSCOCO.}
\vspace{-6mm}
\begin{center}
\small
\setlength{\tabcolsep}{3pt}
\begin{tabular}{l| c |c |c |c |c |c }
\hline
{Methods}  & {{B@1}} & {{B@2}} & {{B@3}} & {{B@4}} & {{M}} & {\textbf{C}} \\
\hline
Graph-Enc-Dec(Avg) & 52.1 & 34.1 & 23.8 & 17.6 & 14.9 & 41.4\\
\hline
Graph-Enc-Dec(Att$^{\ast}$) & 54.3 & 37.0 & 26.8 & 20.3 & 15.9 & 47.2\\
\hline
Graph-Enc-Dec(Att) & 56.0 & 33.6 & 20.1 & 11.9 & 17.0 & 48.5\\
\hline
\end{tabular}
\end{center}	\label{tab:txt_ssg_decoder_unpaired_nogan}
\vspace{-8mm}
\end{table}

\noindent{\textbf{Investigation on Unpaired Setting without GAN.}}
Table~\ref{tab:txt_ssg_decoder_unpaired_nogan} shows the comparisons among different baselines when no explicit cross-modal mapping of the features is done. By feeding the image scene graph directly to the trained scene graph encoder and the sentence decoder, we can  achieve promising performance on the test set. Graph-Enc-Dec (Att) still achieves the best performance in all metrics. This is reasonable since both scene graphs and captions are high-level understandings of the image, and by capturing rich semantic information about objects and their relationships, scene graphs provide an effective way to connect an image to its natural language description. This finding also validates the feasibility of our approach to unpaired image captioning through the use of scene graphs. However, compared to the paired setup (see Table \ref{tab:stoa_coco}), these results are still inferior, meaning that only scene graph is not enough to achieve comparable performance.

\begin{figure*}[t!]
    \centering
    \small
    \vspace{-3mm}
    \includegraphics[width=\linewidth]{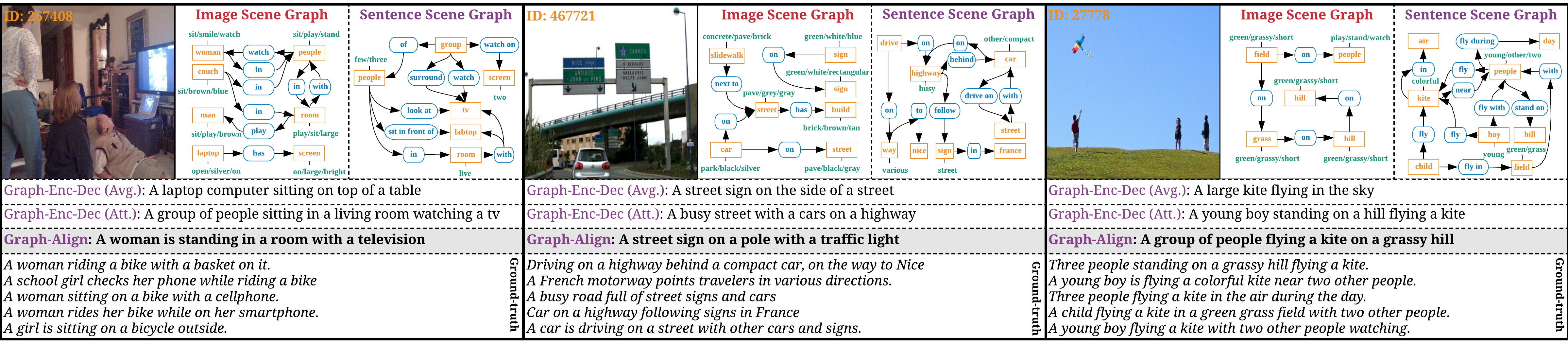}
    \vspace{-6mm}
    \caption{Qualitative examples of different methods. In each example, the left image is the original input image; the middle is the image scene graph; the right image is the ground-truth sentence scene graph for compassion.}
    \label{fig:vis_results_caps}
    \vspace{-3mm}
\end{figure*}

\noindent{\textbf{Investigation on Unpaired Setting with GAN.}}
To align the features from the image modality to the text modality, we use CycleGAN with our Graph-Enc-Dec(Att) model. Table~\ref{tab:coco_gans} shows the comparisons of three kinds of GAN loss: binary \textcolor{\me}{cross-entropy} (BCE) loss with logits (the vanilla GAN loss \cite{goodfellow2014generative}), mean squared error (MSE) loss,  and gradient penalty (GP)~\cite{gulrajani2017improved}. We also compare the results for using different output dimensions in the discriminator.\footnote{For example, for a dimension of 64, the output is a 64-dimensional vector, which is compared against an all-one vector of length 64 for a `Real' input, and with an all-zero vector of length 64 for a `Fake' input.}

\begin{table}
\small
\vspace{-2mm}
\caption{Ablation studies of different GAN losses for Graph-Align model.}
\vspace{-6mm}
\begin{center}
\setlength{\tabcolsep}{4pt}
\begin{tabular}{l l| c |c |c |c |c }
\hline
GAN Loss & Discriminator & {{B@1}} & {{B@2}} & {{B@3}} & {{B@4}} & {{C}} \\
\hline
\multirow{4}{*}{BCE}  & $d_f\rightarrow d_f$ & 64.9  & 44.2 &28.6  & 18.1 & 63.0\\
&  $d_f\rightarrow 64$  & 66.0 & 46.0 & 30.3 & 19.7 & 65.5\\
&  $d_f\rightarrow 1$& 65.5 & 45.4 & 29.6 & 18.8 & 65.2\\
\hline
\multirow{4}{*}{MSE}  & $d_f\rightarrow d_f$  & 65.3 & 44.8 & 28.9 & 18.3 & 62.9\\
&  $d_f\rightarrow 64$  & 66.0 & 45.9 & 29.7 & 18.8 & 63.8\\
&  $d_f\rightarrow 1$& 58.4  & 36.3 & 21.7 & 12.6 & 46.7\\
\hline
\multirow{4}{*}{GP}   &  $d_f\rightarrow d_f$  & 66.1  &46.1  & 30.3 & 19.5 & 65.5 \\
&  $d_f\rightarrow 64$  & \textbf{67.1} & \textbf{47.8} & \textbf{32.3} & \textbf{21.5} & \textbf{69.5}\\
&  $d_f\rightarrow 1$ & 64.5  & 44.2 & 28.5 & 17.9 & 61.1\\
\hline
\end{tabular}
\end{center}\label{tab:coco_gans}
\vspace{-6mm}
\end{table}

\begin{table}[ht]
\small
\caption{\textcolor{\me}{The performances of using different feature mappings on MSCOCO test split. Shared GAN learns a shared feature mapping for three sets of features with CycleGAN. Single GAN concatenates the three kinds of embeddings together and learns a mapping with CycleGAN.}}
\begin{center}
\vspace{-6mm}
\setlength{\tabcolsep}{3pt}
\begin{tabular}{l| r |r |r |r |r |r}
\hline
{Methods}  & B@1 & B@2 & B@3 & B@4 & M & C \\
\hline
Shared GAN & 60.7 & 41.3 & 26.9 & 17.6 & 20.0 & 60.1 \\
Single GAN & 61.8 & 42.1 & 27.3 & 17.7  & 20.1 & 61.2 \\
\hline
Graph-Align & \textbf{67.1} & \textbf{47.8} & \textbf{32.3} & \textbf{21.5} & \textbf{20.9} & \textbf{69.5} \\
\hline
\end{tabular}
\end{center}
\label{tab:different_gans}
\vspace{-8mm}
\end{table}

\begin{table*}[ht]
\caption{Performance comparisons on the test split of the MSCOCO dataset.}
\vspace{-6mm}
\begin{center}
\setlength{\tabcolsep}{4pt}
\begin{tabular}{l| c |c |c |c |c |c |c|c}
\hline
{Method}  & {{BLEU-1}} & {{BLEU-2}} & {{BLEU-3}} & {{BLEU-4}}& {{METEOR}}  & {{ROUGE}} & {{CIDEr}} & {{SPICE}} \\
\hline
\multicolumn{9}{c}{\textit{{Paired Setting}}}\\
\hline
Soft-Attention~\cite{xu2015show} & 71.8 & 50.4 & 35.7 & 25.0 & 23.0 & -- & 90.0 & --\\
Stack-Cap~\cite{gu2017stack} & 78.6 & 62.5 & 47.9 & 36.1 & 27.4 & 56.9 & 120.4& 20.9\\
SGAE (base)~\cite{yang2018shuffle} & 79.9 & -- & -- & 36.8 & 27.7 & 57.0& 120.6 & 20.9 \\
\hline
\multicolumn{9}{c}{\textit{{Unpaired Setting}}}\\
\hline
Language Pivoting~\cite{gu2018unpaired} & 46.2 & 24.0 & 11.2 & 5.4 & 13.2  & --& 17.7\\
\hline
Adversarial+Reconstruction~\cite{feng2018unsupervised} & 58.9 & 40.3 & 27.0 & 18.6 & 17.9 & 43.1& 54.9 &11.1\\
\hline
Graph-Align & \textbf{67.1} & \textbf{47.8} & \textbf{32.3} & \textbf{21.5} & \textbf{20.9} & \textbf{47.2} & \textbf{69.5} & \textbf{15.0}\\
\hline
\end{tabular}
\end{center}
\label{tab:stoa_coco}
\vspace{-8mm}
\end{table*}

We can see that most of the CycleGAN variants improve the performance substantially compared to the results in Table \ref{tab:txt_ssg_decoder_unpaired_nogan}. The GP with 64-dimension discriminator output achieves the best performance. Note that, when we set the output dimension to 1, the performance drops. This indicates that a strong discriminator is crucial for unpaired feature alignments. From the bottom row of Figure~\ref{fig:vis_feature}, we can see that with the help of the mapping module, the three kinds of embeddings are aligned very well, especially the attribute embedding (Figure~\ref{fig:pre_tsne_atr_gan}). It is also worth noting that the triplet features in Figure \ref{fig:pre_tsne_Y_map} are better aligned compared to the raw triplet features in Figure~\ref{fig:pre_tsne_N_map}.

\textcolor{\me}{To further demonstrate the effectiveness of the proposed three feature mapping functions, we conduct additional experiments in Table~\ref{tab:different_gans}. It can be seen that treating the three set of embeddings ($\mathcal{X}_o^k$, $\mathcal{X}_r^k$, and $\mathcal{X}_a^k$) without distinction performs worse than Graph-Align.}

Finally, Table~\ref{tab:stoa_coco} compares the results of the Graph-Align model with those of the existing unpaired image captioning methods \cite{gu2018unpaired,feng2018unsupervised} on the MSCOCO test split. We can notice that our proposed Graph-Align achieves the best performance in all metrics. This demonstrates the effectiveness of our scene graph-based unpaired image captioning model.

\begin{figure}[ht]
    \small
    \vspace{-3mm}
	\centering     
	\includegraphics[width=\linewidth]{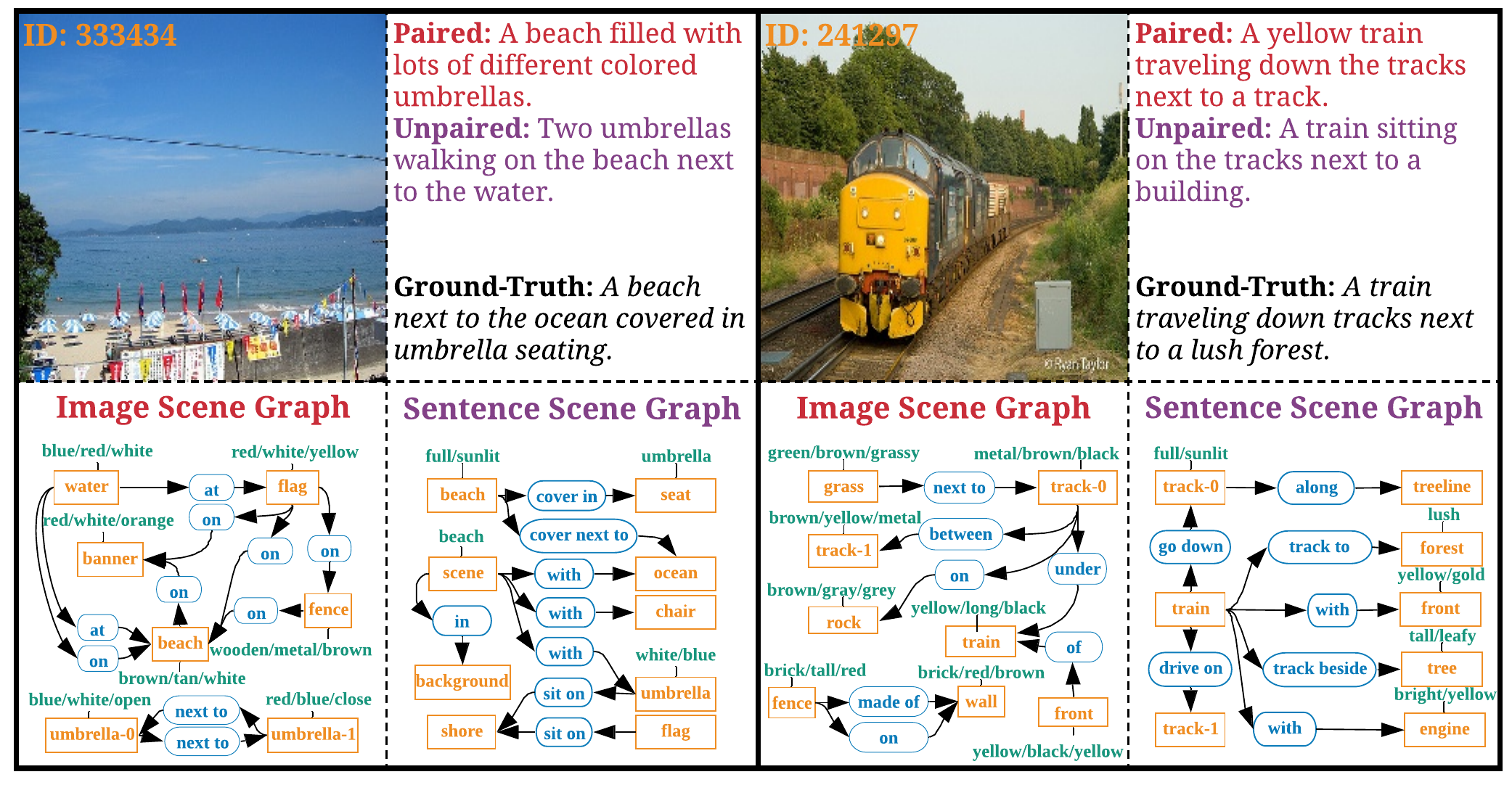}
	\vspace{-6mm}
	\caption{\textcolor{\me}{Examples of unpaired image captioning failure cases. Although the accuracy of image scene graph highly influences the performance of captioning results, our Graph-Align can still generate relevant image captions.}}
	\label{fig:rebuttal_bad_eg}
    \vspace{-3mm}
\end{figure}

\subsection{Qualitative Results}
Figure~\ref{fig:vis_results_caps} visualizes some examples of our models. We show the generated image descriptions using different models along with the ground-truth captions (bottom part).
In the generated image and sentence scene graphs, we mark object, relation, attribute nodes in orange, blue, and green, respectively.
From these exemplary results, we observe that our method can generate reasonable image descriptions by aligning the unpaired visual-textual modalities with the help of scene graphs. Also, we observe that the number of attributes (words in green) in the sentence scene graph is less than that in the image scene graph. This observation potentially explains why there is a huge feature embedding gap between image and text in Figure~\ref{fig:pre_tsne_atr}.

\textcolor{\me}{Figure~\ref{fig:rebuttal_bad_eg} presents some failure cases of our Graph-Align model. We can see that the image scene graphs mainly focus on local regions/objects, while sentence scene graphs convey more information about the images. Such information misalignment leads to generating different captions.}

\section{Conclusions}
In this paper, we have proposed a novel framework to train an image captioning model in an unsupervised manner without using any paired image-sentence data. Our method uses scene graph as an intermediate representation of the image and the sentence, and maps the scene graphs in their feature space through cycle-consistent adversarial training. We used graph convolution and attention methods to encode the objects, their attributes, their relationships in a  scene graph. Our experimental results based on quantitative and qualitative evaluations show the effectiveness of our method in generating meaningful captions, which also outperforms existing methods by a good margin. In future, we would like to evaluate our method on other datasets and explore other mapping methods such as optimal transport. 

\section*{Acknowledgments}
This work was supported in part by NTU-IGS, NTU-Alibaba Lab, NTU DSAIR Center, and NTU ROSE Lab.

\newpage
\bibliographystyle{ieee}
\bibliography{egbib}

\end{document}